\documentclass[journal]{IEEEtran}
\usepackage{amssymb,amsmath}
\usepackage{cite}
\usepackage{graphicx}
\usepackage{psfrag}
\usepackage{url}
\usepackage[latin1]{inputenc}
\usepackage[absolute,overlay]{textpos}
\usepackage{gensymb}
\usepackage{cases}
\usepackage{graphicx}
\usepackage[font=footnotesize]{caption}
\usepackage[font=footnotesize]{subcaption}
\usepackage{float}
\usepackage[linesnumbered,ruled,lined]{algorithm2e}
\usepackage{hyperref}

\usepackage{pgf}

\usepackage[nocomma]{optidef}

\usepackage{amsthm}

\usepackage{booktabs}
\usepackage{color,soul}

\DeclareMathOperator*{\argmax}{arg\,max}

\sethlcolor{green!70!white}

\begin{document}

\title{Traffic Learning and Proactive UAV Trajectory Planning for Data Uplink in Markovian IoT Models}
\author{
    \IEEEauthorblockN{Eslam Eldeeb, Mohammad Shehab, and Hirley Alves}
	\thanks{The authors are with Centre for Wireless Communications (CWC), University of Oulu, Finland. Email: firstname.lastname@oulu.fi (Corresponding author: Mohammad Shehab, email: mohammad.shehab@oulu.fi).} 
	\thanks{This research was supported by the Research Council of Finland (former Academy of Finland) 6G Flagship Programme (Grant Number: 346208). This work has been partly funded by the European Commission through the Hexa-X-II (GA no. 101095759)}
}
\maketitle

\vspace{0mm}
\begin{abstract}

The age of information (AoI) is used to measure the freshness of the data. In IoT networks, the traditional resource management schemes rely on a message exchange between the devices and the base station (BS) before communication which causes high AoI, high energy consumption, and low reliability. Unmanned aerial vehicles (UAVs) as flying BSs have many advantages in minimizing the AoI, energy-saving, and throughput improvement. In this paper, we present a novel learning-based framework that estimates the traffic arrival of IoT devices based on Markovian events. The learning proceeds to optimize the trajectory of multiple UAVs and their scheduling policy. First, the BS predicts the future traffic of the devices. We compare two traffic predictors: \textit{1) the forward algorithm (FA)} and \textit{2) the long short-term memory (LSTM)}. Afterward, we propose a deep reinforcement learning (DRL) approach to optimize the optimal policy of each UAV. Finally, we manipulate the optimum reward function for the proposed DRL approach. Simulation results show that the proposed algorithm outperforms the random-walk (RW) baseline model regarding the AoI, scheduling accuracy, and transmission power.

\end{abstract}

\section{Introduction}
The recent progress in machine-type communication (MTC) based IoT has witnessed new service modes, such as massive MTC (mMTC) and ultra-reliable low latency communication (URLLC)~\cite{6g}. These new service modes introduce critical applications, e.g., remote surgery and vehicle-to-vehicle (V2V) communications, and have raised the need for new technologies to meet the quality-of-service (QoS) demands of such applications~\cite{9083794}. One of the recent critical QoS requirements is the end-to-end latency and real-time data collection~\cite{article}.

Meanwhile, MTC devices (MTDs) have different traffic characteristics compared to traditional human-type communication (HTC) devices~\cite{9504554}. For instance, the MTC packets are usually shorter in length than the HTC packets. Moreover, MTC traffic is highly correlated and more homogeneous, i.e., nearby devices tend to have similar traffic~\cite{6629817}. For example, assume sensor $a$ monitors the package count in an industrial factory and sensor $b$ detects package defects. Let event $1$ correspond to no missing package, whereas event $2$ correspond to a ruined package. In this example, sensor $a$ only is active along with $1$, whereas both sensors will be activated with $2$. It is important to forecast the activation pattern of each sensor to allocate the available resources efficiently. 

In addition, an IoT network can have a massive number of devices within a small area, i.e., mMTC, and it also may include devices with strict reliability and latency demands, i.e., URLLC~\cite{7805265}. The traditional access protocols, such as random access (RA) in LTE and grant-free (GF) non-orthogonal multiple access (NOMA) are efficient in serving HTC; however, they have many limitations operating in IoT networks~\cite{6678832}. For instance, the RA relies on a four-handshake procedure, where the device requests a transmission and waits for the response and identification, which introduces high signaling overhead. transmission, many messages are exchanged between the devices and the base station (BS), which introduces high signaling overhead. In addition, this high signaling overhead, which is necessary for device scheduling, reduces the spectral efficiency of the resources. Moreover, operating on massive deployment, the traditional resource allocation techniques can suffer from a high collision. The high signaling overhead and the collision are considered sources of latency and higher energy consumption, which fails to meet the QoS requirements of the MTC use cases. Other conventional schemes, such as time-division multiple-access (TDMA), also fail to meet the QoS demands of IoT devices~\cite{9845353}. Despite the fairness of the TDMA in scheduling the resources among the devices, it still disrupts the distinction between the active and silent devices, which reduces the spectral efficiency due to the waste of resources.

Machine learning schemes have recently been considered as potential solutions to the aforementioned problems. The learning-based resource allocation schemes aim to reduce the signaling overhead and the collision, consequently reducing the resultant latency. In~\cite{3gpp_fug}, the 3GPP defines the fast uplink as the potential replacement for the traditional RA. It requires a traffic estimation to predict the active and silent devices at a time.
This learning-based access scheme has shown promising results regarding access delay, reliability, collision, and signaling overhead compared to the RA schemes\cite{8663999,8712527}. According to~\cite{9504554}, the initial step in the fast uplink solution is traffic prediction to identify active and silent devices at a time, i.e., a time-series binary classification problem. 

Machine learning tools are efficient in solving time-series problems. Classical machine learning has been widely used in forecasting traffic, such as auto-regressive integrated moving average (ARIMA)~\cite{KHASHEI2009956}, the forward algorithm (FA) in hidden Markov models (HMMs)~\cite{HMM}, and Gaussian mixture models (GMMs)~\cite{GMM}. 
Classical methods rely on statistical equations and probabilistic models to estimate the probability of a device being active or silent at each time instant. They are usually fast and simple with low dimensional problems but suffer from poor accuracy with complex-high dimensional problems. On the other hand, modern machine learning techniques have evolved recently in the time-series problems with the exclusive usage of deep neural networks, such as recurrent neural networks (RNNs), long short-term memory (LSTM), and attention mechanisms~\cite{vaswani2017attention}. Modern deep learning methods rely on feature extraction automatically through deep hidden connected neural network layers. They are normally complex and their training is time-consuming compared to classical techniques. However, they prove to perform better in high-complexity setups.

An important metric that measures the degree of freshness of information received from devices is the age of information (AoI). AoI is defined in~\cite{6195689} as the time difference between the current and generation times of each device's last received packet. It was developed to evaluate the freshness of the data collected from each device\cite{8187436}, where lower AoI means fresher information. There is a direct relation between the traffic prediction and the AoI, where active devices that successfully receive resources and transmit their packets have lower AoI. In contrast, wrong allocation increases the AoI of non-served active devices. In addition, granting the resources to the active devices without receiving transmission requests reduces the signaling overhead, leading to lower latency and lower average transmission AoI of the information. One of the most emerging technologies considered as potential solutions to minimize the AoI future wireless communications is the Unmanned aerial vehicles (UAVs)~\cite{8660516}. Relying on their flexibility, accessibility, and ability to be fully controllable~\cite{8641422}, deploying UAVs enables dynamic and real-time data collection, allowing critical applications to operate safely~\cite{8813065}. Introducing the UAV as a flying BS has many advantages, such as enhancing the line-of-sight (LoS) communication between the device and the UAV as the UAV flies near the served device, improving the throughput, decreasing the transmission energy, enabling the deployment of a massive number of devices in a network, and minimizing the AoI in wireless networks~\cite{book_UAV}. Despite having many advantages, using the UAV as a flying BS has also raised many challenges recently, such as trajectory optimization, flight energy minimization, the freshness of the collected data, and scheduling the resources efficiently among the served devices~\cite{7572034}.

To this end, minimizing the AoI in UAV-based networks is necessary to guarantee that fresh information is received from each device and boost fairness among the devices. Moreover, most of those devices are considered limited-power devices~\cite{9546792}, where the transmission power of the devices needs to be minimized. Therefore, three crucial aspects should be monitored in the UAV-based networks, namely, the regret that describes the accuracy of scheduling the resources, the AoI that exploits the freshness of the information, and the transmission power of the limited-power devices~\cite{8316776}.


\subsection{Related Literature}

Herein, we summarize the existing literature covering the solutions to the traditional RA protocols and the work done on UAV-based networks. To begin with, the authors in~\cite{9119119} proposed a reinforcement learning model to schedule the MTDs to the RA slots, whereas Ali \textit{et al.}~\cite{9075198} exploited the sleeping multi-armed bandit to formulate a fast uplink grant algorithm that prioritizes the device according to their activation and their importance. Their proposed model enhances fairness in the system and decreases the average access delay. In addition, several solutions were mentioned in \cite{6807949}, such as access class barring schemes, dynamic resource allocation, slotted access, and pull-based schemes. However, they still suffer from undesired latency or collision~\cite{8712527}. In~\cite{8417634}, Zhiyi \textit{et al.} overcame the high signaling overhead by introducing a hybrid resource allocation scheduler. The authors in~\cite{Federated_Learning} presented a federated-learning solution to estimate the future traffic of the MTDs. Their work does not consider the latency and complexity analysis that is crucial in designing resource allocation schemes.
 
 
In~\cite{7510870,7881122,8329013}, the optimum number of devices and their positions were optimized to ensure a UAV-based network's capacity or throughput constraints. For instance, the authors in~\cite{7510870} proposed a transport theory-based solution to determine cell boundaries and maximize transmission rate. Meanwhile,~\cite{7881122} used a heuristic algorithm to determine the optimum number of the UAVs and their positions, where their simulation results showed that all the users were meeting their QoS constraints. In addition, \cite{8329013} optimizes the UAV's trajectory to maximize the throughput from the users' perspective.

The works in~\cite{tong2020deep,zhou2019deep, deep_us, deep_china} discussed the AoI minimization via deep reinforcement learning (DRL) solutions. Tong \textit{et al.} presented a trajectory optimization to minimize the AoI while ensuring that the packet drop rate is as low as possible \cite{tong2020deep}. In~\cite{zhou2019deep}, a Markov decision process (MDP) was proposed to formulate the trajectory optimization problem. The authors in~\cite{deep_us,deep_china} formulated a DRL solution to minimize the weighted-sum AoI, where their solutions outperform the baseline schemes such as random walk and distance-based approach. The optimal position of the users was optimized in~\cite{9220821} using a weighted expectation maximization approach. 

Due to the complexity of the UAV-based optimization problems, most of the literature neglected the traffic arrival of the MTDs and assumed them to be active all the time. This assumption is not realistic and completely avoids the resource management aspect of the UAV acting as a flying BS. However, a few recent works have addressed the resource management problem in UAV optimization. For instance,~\cite{9204738} utilized a block successive upper-bound minimization algorithm to jointly minimize the energy consumption and the resource management of the UAV. Moreover, Peng and Shen~\cite{9254093} presented a multi-agent deep deterministic policy gradient solution to the aforementioned problem. They claim that using multiple agents outperforms the single-agent scenario regarding delay and QoS satisfaction ratios.

\subsection{Contribution}
This work addresses the problem of how to exploit a predictive dynamic traffic pattern in order to proactively and efficiently design a UAV trajectory to "navigate and collect data" from IoT devices. Our scheme aims at the jointly minimization of the average AoI of the IoT devices as well as their average transmit power. The proposed algorithm comprises two stages: traffic estimation and UAV learning. Our main contributions are
\begin{itemize}
    \item We design the system model with the aid of the hidden Markov models (HMMs), where Markovian events govern the activation of devices. We assume that multiple UAVs are serving the devices. 
    
    \item We present the FA as the classical traffic estimation approach that estimates device activation probabilities. In addition, we propose an LSTM architecture as the modern deep-learning traffic estimation approach. Both traffic estimators are evaluated in terms of accuracy and complexity. 
    
    \item We propose a DRL solution that optimizes the trajectory path of each UAV and the scheduling policy that jointly minimizes the average AoI, scheduling regret, and average IoT transmission power. Moreover, we acquire the optimum reward function for various devices that yields the best joint performance regarding AoI, scheduling regret, and transmission power.

    \item Exploiting this, the DRL-based UAV trajectory and scheduling optimization outperforms the baseline random-walk (RW) scheme. Our simulation results show that the performance of the proposed algorithms approaches the genie-aided case (the one that uses true activation probabilities instead of the predicted ones). The LSTM traffic predictor shows better AoI and transmission power results than the FA traffic predictor.

\end{itemize}

\subsection{Outline}
The rest of the paper is organized as follows: Section~\ref{SYS_MODEL_SEC} illustrates the system model and the problem formulation. Section~\ref{TRAFF_EST_SEC} discusses the traffic estimation stage, whereas Section~\ref{DQN_SEC} presents the UAV learning stage and the proposed DQN solution. The key performance indicators are described in Section~\ref{EVAL_SEC}. Section~\ref{RES_SEC} depicts the numerical results, and Section~\ref{CONC_SEC} concludes the paper.

\section{System Model}\label{SYS_MODEL_SEC}
Consider an uplink model of $D$ static IoT devices, where$\mathcal{D}=\{1,2,\cdots,D\}$. The devices are randomly distributed in a grid world and served by a set $\mathcal{U} =\{1,2,\cdots, U\}$ of $U$ rotary-wing UAVs that fly with fixed velocity $v_u$ at height $h_u$ and transmit the collected information from the IoT devices to a static BS of height $h_{BS}$. The location of a device $d$ is given by $l_d = (x_d,y_d)$, while the location of the UAV is projected on the 2D plane as $l_u = (x_u,y_u)$, and the BS is located at the center of the grid world, where $l_{BS} = (0,0)$.

\begin{figure}[t!]
	\centering
	\includegraphics[width=0.99\columnwidth]{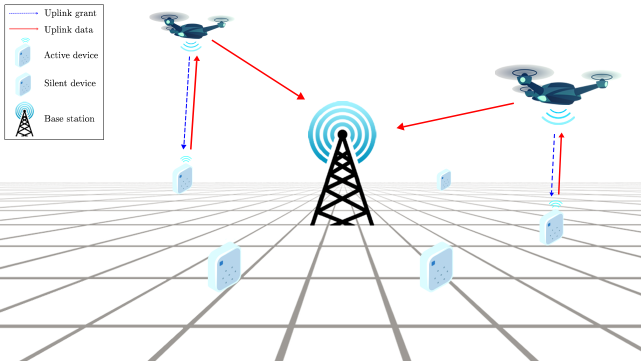}
	\caption{The system model: IoT devices are served by multiple UAVs that relay the information to the BS located at the center of the grid world.} \vspace{2mm}
	\label{Fig1}
\end{figure}

The distance between an IoT device $d$ and a serving UAV $u$ is denoted by $L_{du}$, while 
the distance between a serving UAV $u$ and the BS is denoted by $L_{uBS}$,  
and the distance between two horizontal or vertical points on the grid world is given by $L_g$. Four charging depots are located at the corners of the grid world to enable the UAVs to recharge. The time axis is discretized into $[\tau, 2\:\tau, ...]$, where $\tau$ is the time needed for the UAV to navigate from one grid point to another, i.e., $\tau = \frac{L_g}{v_u}$. During the time $\tau$, the UAV can allocate a resource to one IoT device $d$ $(\alpha_d(t)=1)$ and the scheduling vector is one-hot vector given by $\boldsymbol{\alpha}(t) = [\alpha_1, \alpha_2, ... \alpha_D]$. The system model is illustrated in Fig.~\ref{Fig1}.

\subsection{System Analysis}
Assuming a LoS\footnote{The non-LoS and UAV variable height cases are straightforward to analyze. However, this work focuses on the joint traffic prediction and trajectory planning problem rather than considering different channels and flight models.} communication between the UAV and the BS, and between the IoT devices and the UAV, the channel gain between the UAV and the BS can be calculated as
\begin{align}
\label{eq_3}
    g_{u,BS}(t) &= g_0L_{u,BS}^{-2} = \frac{g_0}{|h_u-h_{BS}|^2+||l_u(t)||^2},
\end{align}
$g_0$ is the channel gain at the reference distance (1 m). Each UAV has a battery of capacity $E$, which is discretized into $e_{max}$ energy quanta. Each energy quanta has energy of $\frac{E}{e_{max}}$. The battery evolution $e_u$ of each UAV can be formulated as~\cite{9815722,9950310}
\begin{equation}
	e_u{(t\!+\!1)}=
	\begin{cases}
		e_u(t)-\lceil e_u^t(t)+e_u^f(v_u) \rceil, & \text{if}   \ \alpha(t) \text{ is non-zero}, \\
		e_u(t)-\lceil e_u^f(v_u)\rceil, & \text{otherwise}, \end{cases}
		\label{available_energy}
\end{equation}
where $\lceil \: \rceil$ is the ceiling approximation, $e_u^t$ is the energy consumed due to the UAV and BS communication, and $e_u^f$ is the energy consumed due to movement. Here, $e_u^t$ can be calculated at time instant $t$ as 
\begin{align}
    e_u^t(t) &= \frac{e_{max}}{E}\:\frac{\sigma^2}{g_{u,BS(t)}}\big(2^{\frac{M}{B}}-1\big),
\end{align}
where $\sigma^2$ is the noise power, $M$ is the packet size of the sensor updates and $B$ is the signal bandwidth. In addition, according to~\cite{9162896}, $e_u^f$ is formulated as 
\begin{align}
e_u^F(v_u) = &\frac{e_{max}}{E}\: \left[P_0 \left( 1+\frac{3v_u^2}{v_{tip}^2} \right)  \right. \nonumber\\
& \left.+P_1\left(\sqrt{1+\frac{v_u^4}{4v_0^4}}-\frac{v_u^2}{2v_0^2}\right)^\frac{1}{2}+\frac{1}{2}v_u^3 d_0\rho \mu_0 Z \right],
\end{align}
where $P_0$ and $P_1$ represent the blade profile power and derived power when the UAVs are hovering, respectively, $v_t$ describes the velocity of the UAVs, and $v_{tip}$ depicts the tip speed of the blade. Meanwhile, $v_0$ is the mean rotor-induced velocity when hovering, $d_0$ represents the fuselage drag radio, $\rho$ is the air density, $\mu_0$ represents the rotor solidity and $Z$ the area of the rotor disk.


\subsection{Traffic Arrival}
We denote the activation of a device $d$ at a time instant $t$ as $w_d(t)$, where $w_d(t) = 1$ means that device $k$ is active at time instant $t$ and $w_d(t) = 0$ means it is silent. Hence, the activation vector of the IoT devices in the network at time instant $t$ can be written as $\boldsymbol{W(t)} = \left\lbrace w_1(t), w_2(t), ..., w_D(t) \right\rbrace$. Consider a set of $\mathcal{K}=\{1,2,\cdots, K\}$ of $K$ events that control the activation of the IoT devices. Each event is considered an event-driven background Markovian On-off process that influences the IoT devices \footnote{We note that many outdoor or indoor events may follow  Markovian behaviour of occurrence. The reader can refer to examples related to indoor fire events and autonomous vehicles in \cite{9845353}, which utilizes the same model for the sake of illustration in predictive traffic scenarios.}. 

\begin{figure}[t!]
	\centering
	\includegraphics[width=0.9\columnwidth]{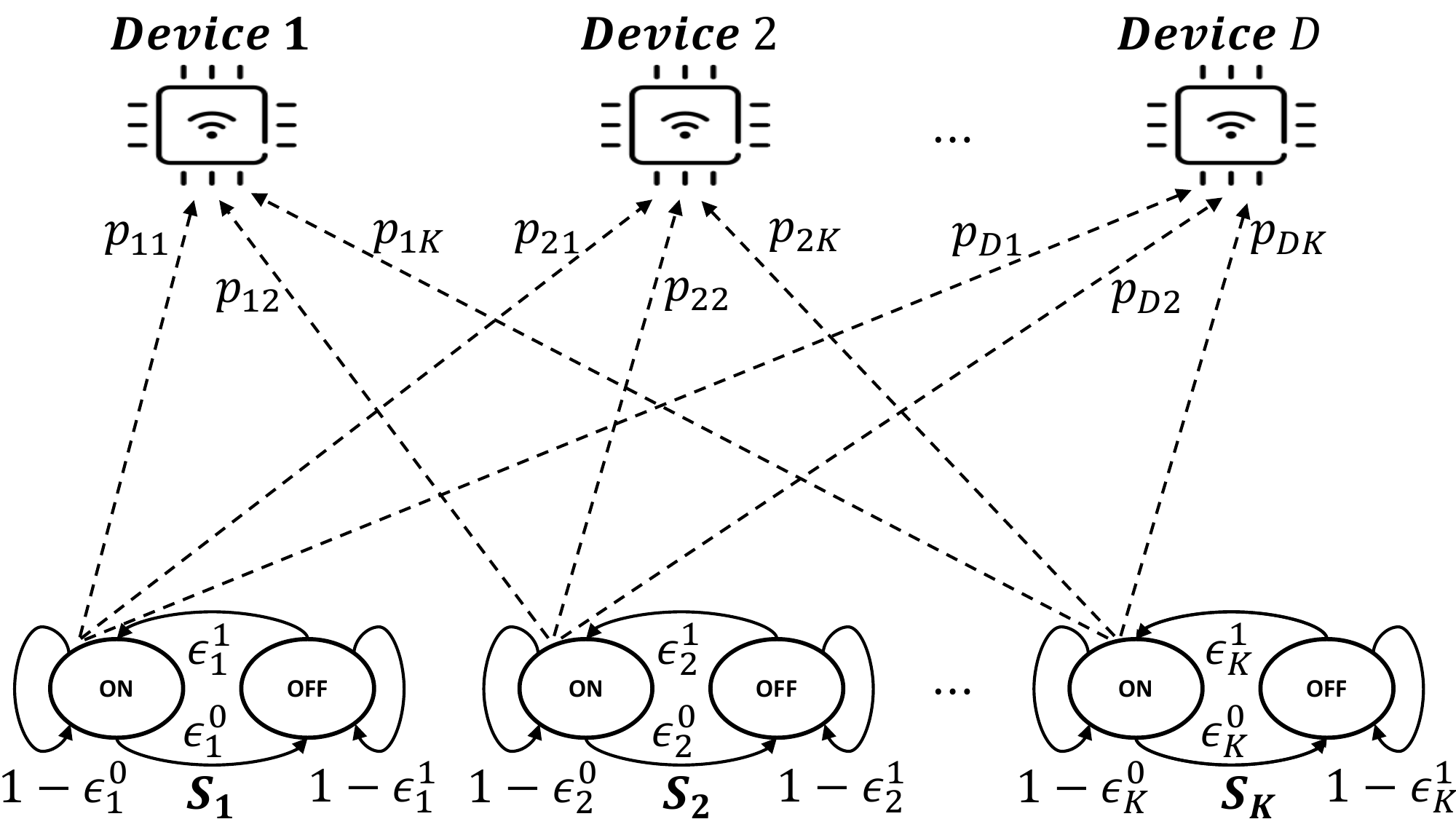}
	\caption{The activation of $D$ devices is modeled as a Markovian arrival of $K$ binary events. If an event $k$ is active, it influences a device $d$ with an activation probability of $p_{dk}$.} 
	\vspace{-0mm}
	\label{Fig2}
\end{figure}

We denote the activation of an event $k$ at time instant $t$ as $S_k(t)$, where $S_k(t) = 1$ means that the event $k$ is in the ON state at time instant $t$ and $S_k(t) = 0$ means its existence in the OFF state. Hence, the activation vector of the binary events in the network at time instant $t$ can be described as $\boldsymbol{S(t)} = \left\lbrace S_1(t), S_2(t), ..., S_K(t) \right\rbrace$. As shown in Fig.~\ref{Fig2}, the activation model is described as a set of binary Markov chains with transition probabilities $\epsilon_k^{(1)}$, which is the transition probability of an event $k$ from on state ($S_k(t)=0$) to off state ($S_k(t+1)=1$) and $\epsilon_k^{(0)}$, which is the transition probability of an event $k$ from on state ($S_k(t)=1$) to off state ($S_k(t+1)=0$). The transition between states for each Markov chain can be summarized as follows
\begin{align}
    \Pr\left(S_k(t+1)=0 \middle| S_k(t)=1\right) &=\epsilon_k^0, \\
    \Pr\left(S_k(t+1)=1 \middle| S_k(t)=1\right) &=1-\epsilon_k^0,\\
    \Pr\left(S_k(t+1)=1 \middle| S_k(t)=0\right) &=\epsilon_k^1, \\
    \Pr\left(S_k(t+1)=0 \middle| S_k(t)=0\right) &=1-\epsilon_k^1,
\end{align}
with the transition matrix
\begin{align}\label{State_Matrix}
\begin{split}
P_{\boldsymbol{S}} = 
\begin{pmatrix}
1-\epsilon_k^1 & \epsilon_k^1 \\
\epsilon_k^0 & 1-\epsilon_k^0 \end{pmatrix}.
\end{split}
\end{align}

Moreover, each active event $\boldsymbol{S}(t) = 1$ has a probability $p_{dk}$ to activate a device $d$ at time instant $t$. In contrast, a silent event $\boldsymbol{S}(t) = 0$ has a zero probability of activating a device $d$ at time instant $t$. Therefore, the probability of a device $d$ to be active or silent affected by an event $k$ at time instant $t$ can be calculated as follows  
\begin{align}
    \Pr\left(w_d(t)=1 \middle| S_k(t)=1\right) &=p_{dk}, \\
    \Pr\left(w_d(t)=0 \middle| S_k(t)=1\right) &=1-p_{dk},\\
    \Pr\left(w_d(t)=1 \middle| S_k(t)=0\right) &=0, \\
    \Pr\left(w_d(t)=0 \middle| S_k(t)=0\right) &=1,
\end{align}
with the activation matrix
\begin{align}\label{State_Matrix_1}
\begin{split}
P_{\boldsymbol{W}} = 
\begin{pmatrix}
p_{1,1} & p_{1,2} & \cdots & p_{1,K}\\
p_{2,1} & p_{2,2} & \cdots & p_{2,K} \\
\vdots & \vdots & \ddots & \vdots \\
p_{D,1} & p_{D,2} & \cdots& p_{D,K}
\end{pmatrix}.
\end{split}
\end{align}
In addition, the activation probability of a device $d$ affected by all the events in the network can be formulated as follows 
\begin{align}\label{activation_Prop}
   \Pr\left( w_d(t)=1 \middle| \boldsymbol{S}(t)\right)&=1-\bigcap_{k=1}^K \Pr\left( w_d(t)=0 \middle| S_k(t)  \right) \\
   &=1-\prod_{k=1}^K (1-p_{dk})^{S_k(t)}.
\end{align}

\subsection{Problem formulation}

We aim to jointly minimize the average AoI, the accumulative regret, and the device's average transmission power. However, before we cast the optimization problem, we introduce the metrics addressed in the problem formulation.
\subsubsection{Age of Information}
The AoI is used to measure the freshness of the transmitted packets and the network fairness among the devices~\cite{bedewy}. The AoI for device $d$ can be formulated as the difference between the current time instant $t$ and the last time slot $t_d$ such that $u_d(t_d)=1$. If a device transmits an update packet at instant $t$, i.e., $\alpha_d(t)=1$, its AoI is reset to one. To reduce the AoI, the UAVs need to forecast the active devices and serve those with longer AoI. We formulate the discrete AoI of device $d$ as follows 
\begin{equation}
\label{AoI_update}
	A_d(t) =
	\begin{cases}
		1, & \quad \text{if} \ \alpha_d(t)=1, \\
		\text{min}\{A_{max},A_{d}(t-1) + 1\}, & \quad \text{otherwise}, 
	\end{cases}
\end{equation}
where $A_{max}$ is the maximum AoI threshold in the model. The average age of the network at time instant $t$ is calculated as
\begin{equation}
    \Bar{A}(t)=\frac{1}{D}\sum_{d=1}^D A_d(t).
\end{equation}

\subsubsection{Accumulative regret}
Regret occurs when allocating a resource to an inactive device while an active device is left unserved~\cite{9845353}. For example, consider a network of two devices $d_1$ and $d_2$, which are active and silent, respectively, i.e., $w_{d_1}(t) = 1$ and $w_{d_2}(t) = 0$. Suppose the scheduling policy is $\alpha = [0,1]$, i.e., $\alpha_{d_1}(t) = 0$ and $\alpha_{d_2}(t) = 1$. The network has scheduled a resource to an inactive device while an unserved active device exists. Therefore, the regret in this scenario is $1$. The regret at time instant $t$ can be computed as the minimum value among wrongfully scheduled resources $\omega_t$ and the missed scheduled resources $\eta_t$
\begin{equation}
    R(t) = \min \left \lbrace \omega_t,\eta_t \right \rbrace,
\end{equation}
and the accumulative regret at time instant $T$ is formulated as
\begin{equation}
    R_c(T)=\sum_{t=1}^T R(t).
\end{equation}

\subsubsection{Transmission power}
The transmission power $P_d$ of an IoT device $d$ at time instant $t$ is calculated as
\begin{align}
    P_d(t) = 
    \left(2^{\frac{M}{B}}-1\right)\frac{\sigma^2}{g_0}\:\left(L_{du}^2(t) + h_u^2\right).
\end{align}
The average transmission power of the network is 
\begin{equation}
    \Bar{P}(t)=\frac{1}{D}\sum_{d=1}^D P_d(t).
\end{equation}

\subsubsection{Joint optimization problem}
We are now ready to cast the joint optimization of the average age of information, the accumulative regret, and the average transmission power as:
\begin{subequations}\label{P1}
	\begin{alignat}{2}
	\mathbf{P1:} \ &\underset{\boldsymbol{\alpha}(t),\boldsymbol{l}(t)}{\min}       &\ \ \ & \frac{1}{T}\sum_{t=1}^T \frac{1}{D} \sum_{d = 1}^{D}A_d(t) + \zeta_1 \: P_d(t) + \sum_{t=1}^T \zeta_2 \: R(t),\label{P1:a}
	\ \\
	& \ \text{s.t.}   &      & \sum_t^{T_u}P_u(v_u)\leq e_u(t), \label{P1:b}\\
		& & & l_u(1) = l_{c,u}, \label{P1:c}
	\end{alignat}
\end{subequations}
where $l_{c,u}$ are the coordinates of the charging depot where UAV $u$ is going to take off. Therefore, each UAV is forced to start its trajectory from one of the corners of the grid world. The constraint in~\eqref{P1:b} is to ensure that the UAV still has enough energy before moving back to the nearest corner to recharge whenever a low battery is monitored. During recharging, the IoT devices transmit their information directly to the BS to overcome long waiting times. In addition, $\zeta_1$ and $\zeta_2$ are priority factors that affect the trade-off between the importance of minimizing the AoI, transmission power, and the regret according to the design requirements.

\section{The Traffic Prediction Stage}\label{TRAFF_EST_SEC}
\begin{figure}[t!]
	\centering
	\includegraphics[width=1\columnwidth]{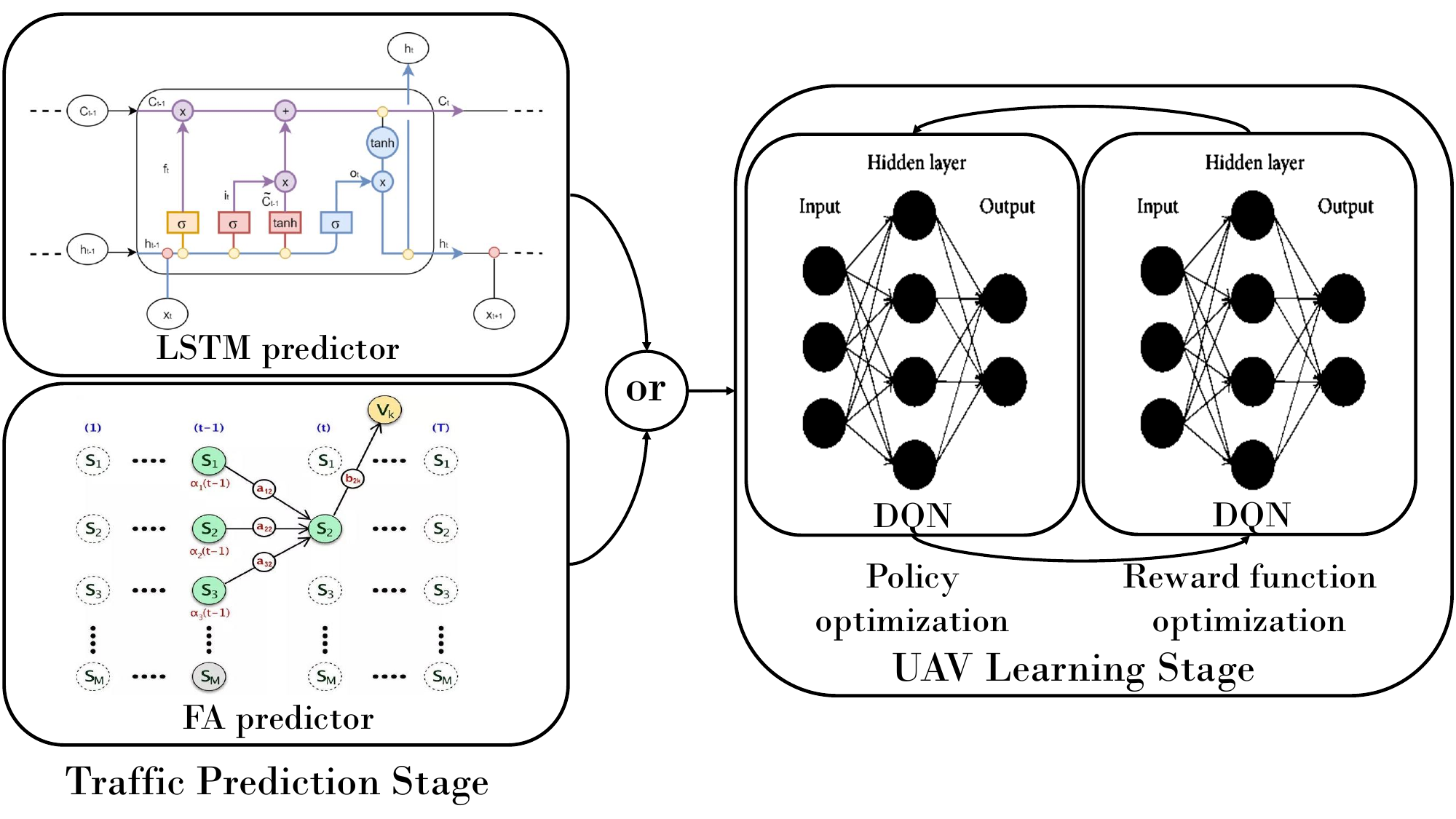} \vspace{2mm}
	\caption{The stages of the proposed algorithm.} 
	\vspace{0mm}
	\label{Fig33}
\end{figure}

As shown in Fig.~\ref{Fig33}, the proposed algorithm has two stages: 
\begin{itemize}
    \item the traffic prediction stage, and 
    \item the UAV learning stage. 
\end{itemize}

\begin{figure}[t!]
	\centering
	\includegraphics[width=0.7\columnwidth]{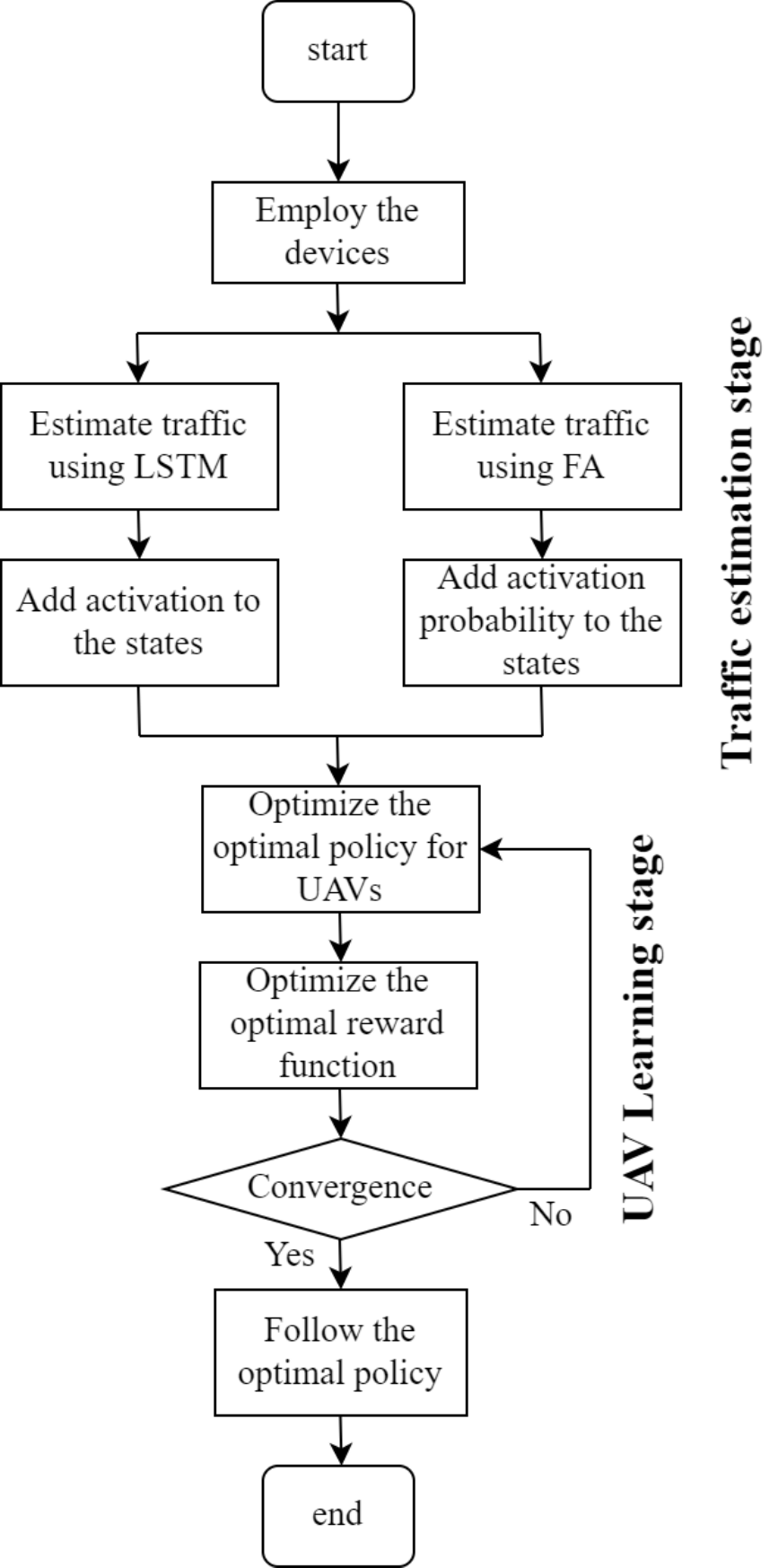} \vspace{2mm}
	\caption{Flow chart of the proposed algorithm. First, the BS estimates the traffic using the LSTM or the FA in the traffic estimation stage. Then, the optimal policy of the UAVs is optimized in the UAV learning stage.}
	\vspace{0mm}
	\label{Fig3}
\end{figure}

In the UAV learning stage, the agents optimize the optimal policy to follow and optimize the values of $\zeta_1$ and $\zeta_2$ in the reward function for different device deployments.
%
Fig.~\ref{Fig3} shows the flowchart of the proposed algorithm and the relationship among the stages.
This section introduces the traffic estimation stage, while the UAV learning stage is discussed in the next section. 

Herein, the objective is to minimize the AoI, regret, and power jointly. However, to minimize regret, the UAVs need to know which devices are active, thus granting them a resource and those which are silent, avoiding wastage of available resources at each time instant.
In addition, there is a correlation between the AoI and regret. For instance, suppose having an active device $d_1$ at time instant $t$, i.e., $w_{d_1}(t) =1$ and a scheduling policy $\alpha_{d_1}(t) = 1$. In this case, the regret is zero, and the AoI of that device is reset to 1. On the other hand, if the scheduling policy of the active device $d_1$ is $\alpha_{d_1}(t) = 0$, the regret is one, and the AoI would also be incremented according to \eqref{AoI_update}. A straight positive correlation between both metrics is not always the case. Therefore, a good traffic predictor is needed for joint optimization. 

In this section, we present an HMM architecture as the proposed classical FA traffic predictor and an LSTM architecture as the potential modern traffic predictor. We compare both architectures from different point-of-views, such as the inputs, the outputs that will be used by the UAV to perform the scheduling, the space complexity, the time complexity, and the accuracy.

\subsection{The Forward Algorithm}
As described in the previous section, the activation of IoT devices is completely affected by the state of the background events. In addition, those states are unknown to the BS (hidden), i.e., the BS does not have information about the active and inactive events. Therefore, we can model the relation between the events and the devices as HMMs~\cite{HMM}. The HMMs consist of a set of unknown events and observations (device activations) affected by the states of those events.
%
%
The activation probability of the devices $Y_d = \Pr\left( w_d(t)=1 \middle| \boldsymbol{S}(t)\right)$ is calculated from \eqref{activation_Prop}. The major concern is that the latter equation relies on the knowledge of the unknown states. The FA can estimate the hidden states by computing the joint distribution between the states and the observations recursively as 
\begin{align}
\label{alpha_equation}
&\Pr\left(\boldsymbol{S}(t),\boldsymbol{W}(1:t)\right) = \Pr \left(\boldsymbol{W}(t) \middle| \boldsymbol{S}(t)\right) \cdot\nonumber\\
&\sum_S^{k(t-1)} \Pr\left(\boldsymbol{S}(t)\middle|\boldsymbol{S}(t-1)\right) \Pr\left(\boldsymbol{S}(t-1),\boldsymbol{W}(1:t-1)\right).
\end{align}
Then the state-activation joint probability is maximized over all possible events using ~\cite{HMM2} 
\begin{equation}\label{eq:most_likely_S}
\boldsymbol{S^{*}}(t)=\argmax_{\boldsymbol{S}(t)} ~ \Pr\left(\boldsymbol{S}(t),\boldsymbol{W}(1:t)\right).
\end{equation}
The BS utilizes the estimation of the hidden states of the events to predict the activation of the IoT devices. The predicted activation probability of device $d$ at time instant $t+1$ given the predicted hidden states $\boldsymbol{S^{*}}(t)$ can be formulated as
\begin{align} \label{activation_t_1}
   \tilde{Y}_d &= \Pr\left(w_d(t+1)=1 \middle| \boldsymbol{S^{*}}(t)\right) \nonumber\\
   &=1-\bigcap_{k=1}^K \Pr\left( w_d(t+1)=0 \middle | \boldsymbol{S^{*}}_k(t)  \right) \\
   &=1-\prod_{k=1}^K \begin{cases}
        1-\epsilon_k^1+\epsilon_k^1(1-p_{dk}), & \quad \boldsymbol{S^{*}}_k(t)=0,\\
       \epsilon_k^0+(1-\epsilon_k^0)(1-p_{dk}), & \quad  S^{*}_k(t)=1.
     \end{cases}
\end{align}
The output of the FA is an estimated probability of each device being active.

\subsection{Long Short-Term Memory}\label{LSTM_Illustration}
The LSTM was introduced in~\cite{article_LSTM} to solve the problem of vanishing gradient in RNNs resulting from long sequences~\cite{279181}. It proposes a short memory $h(t)$ for short series in the past and a long memory $C(t)$ to store the relevant information from the long sequences. As shown in Fig. \ref{Fig33}, the LSTM consists of 4 gates:
\begin{enumerate}
    \item The forget gate $f(t)$: it is used to extract the relevant information from the input to be stored in the long memory and forget irrelevant information. Its updated equation is
    \begin{equation}
    f(t) = \sigma (w_f \, [h(t-1),x(t)]),
    \end{equation}
    where $\sigma$ is a sigmoid activation function, $w_f$ are the weights to be updated, $h(t-1)$ is the previous hidden layer, and $x(t)$ is the input features.

    \item The learn gate $i(t)$: it works similarly to RNN as its main purpose is to learn new patterns from the short sequences. Its updated equations are
    \begin{align}\label{Lear_Gate}
            i(t) &= \sigma (w_i \, [h(t-1),x(t)]), \\
            \tilde{C}(t) &= \tanh \: \left(w_c \, [ h(t-1),x(t) ] \right),
    \end{align}
    where $\tilde{C}(t)$is a vector of the potential features to be used to update the long memory.
    
    \item The remember gate: it uses the result of the forget gate and the learn gate to update the long memory, where
    \begin{equation}\label{Remember_Gate}
        C(t) = f(t) \: C(t-1) + i(t)  \: \tilde{C}(t).
    \end{equation}
    
    \item The use gate: it updates the short memory as
    \begin{align}\label{Use_Gate}
        o(t) &= \sigma (w_o[h(t-1),x(t)]), \\
        h(t) &= o(t) \: \tanh(C(t)).
    \end{align}
    
\end{enumerate}

One of the strongest aspects of modern deep learning techniques in time-series prediction is that they rely on observations only. Therefore, the LSTM only needs to collect a sufficient amount of data generated by the described model that can efficiently describe the model hyperparameters and the hidden states. It uses the collected observations from history and captures their pattern to estimate possible future observations. The output of the LSTM is binary as it returns which devices are expected to be active or silent at each time instant.

The output of this stage is a vector that describes the activation of the devices. Its size is the number of devices in the network $D$. This vector is either an activation probability vector $\in \left[0,1\right]$, in case of using the FA as the traffic predictor, or a binary vector, in case of using the LSTM as the traffic predictor.


\section{The UAV Learning Stage}\label{DQN_SEC}
In this section, we formulate the addressed problem as an MDP. A DRL-based solution is presented, where we cast the reward function that will be used in the UAV learning stage to jointly minimize the average AoI, transmission power, and accumulative regret. 
\subsection{Markov Decision Process}
An MDP is usually described in terms of the tuple $\langle s, a, r, p \rangle$, which consists of the state $s$, the action $a$, the reward $r$, and the state transition probability $p$. In addition, the environment is the IoT network modeled in Section~\ref{SYS_MODEL_SEC} and the agents are the UAVs that serve the devices. At time instant $t$, the agents are found at state $s(t)$ and select an action $a(t)$. Each agent moves to a new state $s(t+1)$ following the state transition probability $p_{a(t)}(s(t),s(t+1))$ of the environment. In addition, the agents gain an immediate reward of $r(t)$ based on the selected action that transits the agent from one state to another. The agents aim to maximize the received reward, which is usually formulated in terms of the desired functions to be minimized or maximized. Here, the reward function will be formulated in terms of the average age, transmission power, and accumulative regret as in \ref{P1}. A policy $\pi$ is the strategy that the agents would follow to select a particular action at each state. Whenever the agent selects an action that results in a state that has low AoI, transmission power, and accumulative regret, that agent will receive a higher reward. Therefore, the agent's task is to discover the best possible action at each state that results in the best possible reward. This process is been referred to as the optimal policy $\pi_*$.

\subsubsection{State space}\label{STATE_SPACE_SECTION}
In the described problem, the state space at time instant $t$ consists of four elements: \textit{i)} the AoI vector $A(t) = [A_1(t), ..., A_D(t)]$, \textit{ii)} a position vector of each UAV $[l_1, ..., l_U]$, \textit{iii)} the parameter $\Delta_u$ that describes the difference between the available energy at each UAV and the required energy to reach the nearest charging depot, and \textit{iv)} the predicted activity vector of each device $\boldsymbol{W}(t)$ in the case of using the LSTM architecture for the traffic prediction or an activation probability vector of each device $[\Pr\left( w_1(t)=1 \middle| \boldsymbol{S}(t)\right), ..., \Pr\left( w_D(t)=1 \middle| \boldsymbol{S}(t)\right)]$ in the case of using the forward algorithm for the traffic prediction. 

\subsubsection{Action space}
The action space at time instant $t$ consists of two elements: \textit{i)} the device to be served by each UAV $\alpha(t) = [\alpha_1, ..., \alpha_U]$ and \textit{ii)} the movement of the UAV $\beta_u(t) = [north, south, east, west, hovering]$.

\subsubsection{State transition probability}
We assume a  deterministic state-space transition probability, thus each component of the state vector is affected by deterministic transition equations. For instance, the AoI is updated according to \eqref{AoI_update}, and the position of each UAV is updated according to
\begin{equation} \label{eqn:directions}
	l_u(t+1)=
	\begin{cases}
		l_u(t)+(0,L_g), & \quad \beta_u(t)=\text{north}, \\
		l_u(t)-(0,L_g), & \quad \beta_u(t)=\text{south}, \\
		l_u(t)+(L_g,0), & \quad \beta_u(t)=\text{east}, \\
		l_u(t)-(L_g,0), & \quad \beta_u(t)=\text{west}, \\
		l_u(t), & \quad \text{Hovering}, \\
	\end{cases}
\end{equation}
and the needed energy before recharge $\Delta_u$ 
is updated by subtracting the difference between the available energy calculated in \eqref{available_energy} and the needed energy to move towards the nearest charging depot. Finally, the traffic predictor outputs the activation pattern or probability.

\subsubsection{Reward function}
Based on the optimization problem in P1, the immediate reward is described as
\begin{equation}
    \label{reward_function}
    r(t) = - \frac{1}{D} \sum_{d = 1}^{D} \left( A_d(t) + \zeta_1 \: P_d(t) \right) - \zeta_2 \: R(t).
\end{equation}
and the accumulative reward is
\begin{equation}
    \label{reward_function}
    r(T) = - \frac{1}{T} \left( \sum_{t=1}^T \frac{1}{D} \sum_{d = 1}^{D} \left(A_d(t) + \zeta_1 \: P_d(t) \right) + \sum_{t=1}^T \zeta_2 \: R(t)\right).
\end{equation}

\subsection{Solving the MDP Problem}
The action-value function $q_{\pi}(s, a)$ is the expected reward starting from a state $s$, taking an action $a$ and then following the policy $\pi$. The optimal action-value function can be described as
\begin{equation}\label{eq:opt_action_value}
q_*(s,a) = \argmax_{\pi} q_{\pi}(s,a),
\end{equation}
where at the optimal policy $\pi_*$, the optimal action-value function is satisfied $q_*(s,a) = q_{\pi_*}(s,a)$. In addition, the optimal policy is simply maximizing, at each state, the action-value function over all the possible actions. The Bellman equation describes the optimal action-value function recursively as
\begin{equation}\label{eq:opt_action_value_2}
q_*(s(t),a(t)) = r(t) + \gamma \sum_{s(t+1)}  \max_{a(t+1)} q_*(s(t+1),a(t+1)),
\end{equation}
where $\gamma \in [0,1]$ is the discount factor that controls how much the agent cares about the future rewards relative to the immediate rewards, i.e, $\gamma = 0$ means that the model cares only about the immediate reward, whereas $\gamma = 1$ means that the model prioritizes the future reward up to infinity. 
The Bellman equation is non-linear and has no closed-form solution. Therefore, iterative solutions are used to solve it. Q-learning is a model-free iterative algorithm that is used to learn how good an action is in a particular state. It is formulated as follows
\begin{align}
& Q\left(s\left(t\right),a\left(t\right)\right) \leftarrow \:  Q\left(s\left(t\right),a\left(t\right)\right) + \nonumber\\
&\alpha \:  \left(r\left(t\right) +  \gamma \: \max_a Q\left(s\left(t+1\right),a\right)
-Q\left(s\left(t\right),a\left(t\right)\right)\right),
\end{align}
where $\alpha$ is the learning rate and $Q(s,a) \rightarrow q_*(s,a)$. The $\epsilon$-greedy policy is used in the Q-learning algorithm such that the model chooses a random action with a probability $\epsilon$  and the greedy action. Thus, such action maximizes the action-value function with probability $1-\epsilon$. Usually, $\epsilon$ is set to be a very large value (close to 1) at the beginning of the learning process and decays with time. This procedure is called the exploration-exploitation trade-off. The larger the value of $\epsilon$, the more the exploration, whereas small $\epsilon$ means that the model is exploiting what it has learned to maximize its action-value function. The Q-learning algorithm is suitable for simple problems, where the state space and the action space are relatively small. However, in high-dimension state and action spaces, such as the described UAV model, Q-learning fails to converge. Therefore, action-value function estimation algorithms are used to solve such problems with high dimensions.

\subsection{The DQN solution}
To overcome the curse of dimensionality of the state and action spaces, DQN was proposed in \cite{DQNs}. The DQN utilizes an artificial neural network (ANN) to estimate the action-value function $Q(s, a|\theta_1)$, where $\theta_1$ is a vector containing the weights of the trained ANN to estimate the action-value function. This ANN is called the estimate network. Therefore, the action-value function is estimated by optimizing the weights that minimize the loss function
\begin{align}
    \mathbf{L}(\theta_1(t)) = &(r(t)+\max_a Q(s(t+1),a|\theta_1(t-1))\\ \nonumber
    &-Q(s(t),a(t)|\theta_1(t)))^2.
\end{align}

The DQN introduces the experience replay and the fixed Q-targets techniques. The experience replay proposes to save the tuple $\langle s, a, r,p \rangle$ in a memory called the replay memory. Then, a mini-batch is sampled from this memory to be used in the training of the estimate network. The fixed Q-targets technique utilizes a new ANN called the target network, where its weights $\theta_2$ are updated every $O$ time instants and are used as the targets for the estimate network. Hence, the loss function is now formulated as
\begin{equation}
    \mathbf{L}(\theta_1) = (r(t)+\max_a Q(s(t+1),a|\theta_2)
    -Q(s(t),a(t)|\theta_1))^2,
\end{equation}
and the weights are optimized using stochastic gradient descent methods. Thus, the weights are updated as follows

\begin{align}
\label{DQN_update}
    \theta_1 = &\theta_1 + \alpha (r(t) + \max_a Q(s(t+1),a|\theta_2) - \\ \nonumber
    &Q(s(t),a(t)|\theta_1)) \: \nabla_{\theta} Q(s(t),a(t)|\theta_1),
\end{align}
where $\nabla_{\theta_1}$ is the gradient with respect to $\theta_1$.

Choosing the values for $\zeta_1$ and $\zeta_2$ in the reward function controls the resulting AoI, regret, and transmission power. Therefore, the BS needs to optimize the best values for $\zeta_1$ and $\zeta_2$ that jointly minimize the AoI, regret, and transmission power to the global minimum given a certain setup. We propose a DQN architecture to estimate the optimal reward function based on the best values for $\zeta_1$ and $\zeta_2$ for a given setup. For the new DQN, the states are the number of devices in the network and the action space consists of the chosen values for $\zeta_1$ and $\zeta_2$. The reward function for the new DQN is simply negative the multiplication of the average AoI, accumulative regret, and the average transmission power calculated from the first trained DQN within an episode of time $T$
\begin{equation}
    \label{reward_function_2}
    r_{DQN} = - \: \Bar{A}(T) \: R(T) \: \Bar{P}(T).
\end{equation}
The weights for the new estimate and target networks are $\theta_3$ and $\theta_4$, respectively. 

The reward function of the new network depends on the reward function of the initially trained network and the reward function of the initial network depends on optimizing the values of $\zeta_1$ and $\zeta_2$ using the second DQN. Therefore, the problem is solved iteratively between both networks. Note that this approach finds the optimized priority factors that jointly minimum AoI, power consumption, and regret; however, these factors could change depending on the application's priority. Algorithm \ref{alg1} summarizes the proposed initial DQN, where the forward algorithm is used as the traffic predictor, whereas Algorithm \ref{alg2} describes the proposed initial DQN, where an LSTM\footnote{It is worth mentioning that deep recurrent Q-networks (DRQN)~\cite{hausknecht2017deep} could be used here, where the recurrent neural network is considered as a part of the DQN itself instead of having it as a separate layer before the DQN. However, we use a separate LSTM layer to have a fair comparison between the output of the FA and the LSTM to the states of the DQN and their influence on the performance of the system.} architecture is used as the traffic predictor. Finally, Algorithm \ref{alg3} presents the reward function optimization\footnote{This algorithm is a centralized DRL algorithm, where the training is performed at the BS; therefore, there is a small message exchange between the UAVs and the BS. For instance, the UAVs send their current states to the BS, whereas the BS transmits the optimized actions to the UAVs.}.

\begin{algorithm}[!t]
\SetAlgoLined
Define the number of devices $D$ and their coordinates $l_d$.

Define $\epsilon$, $\gamma$, $\alpha$ and $O$.

Estimate the hidden states using \eqref{eq:most_likely_S}.

Calculate the device's activation probabilities using \eqref{activation_t_1}.

Utilize the estimated probabilities from~\eqref{activation_t_1} in the state space.

Define the reward function in \eqref{reward_function}.

Train the DQN in algorithm \ref{alg3} to optimize $\zeta_1$ and $\zeta_2$.

Define the number of episodes $E$.

Initialize $t=1$.

\For{e = 1,...,$E$}{
    \While{$\Delta_u(t) > 0$}{
        Choose a random action $a$ with probability $\epsilon$ or select the greedy action $a = \max_a Q(s(t),a)$ with probability $1-\epsilon$.
        
        Save $\langle s(t),a(t),r(t),p(t) \rangle$ in the replay buffer.
        
        Sample a mini-batch from the buffer.
        
        Update $\theta_1$  and $\theta_2$ every $O$ instants using~\eqref{DQN_update}.
        
        $t = t+1$.
    }
}

\caption{The proposed DRL algorithm with the FA as the traffic predictor.}
\label{alg1}
\end{algorithm}

\begin{algorithm}[!t]
\SetAlgoLined
Define the number of devices $D$ and their coordinates $l_d$.

Define $\epsilon$, $\gamma$, $\alpha$ and $O$.

Generate an activation sequence for each device $\boldsymbol{W(t)}$ at each instant $t$.

Use an LSTM to predict the future activation using the past sequence as illustrated in Section~\ref{LSTM_Illustration}.

Utilize the predicted activity for each device $w_D(t)$ in the state space as illustrated in Section~\ref{STATE_SPACE_SECTION}.

Define the reward function in \eqref{reward_function}.

Train the DQN in Algorithm \ref{alg3} to optimize $\zeta_1$ and $\zeta_2$.

Define the number of episodes $E$.

Initialize $t=1$.

\For{e = 1,...,$E$}{
    \While{$\Delta_u(t) > 0$}{
        Choose a random action $a$ with probability $\epsilon$ or select the greedy action $a = \max_a Q(s(t),a)$ with probability $1-\epsilon$.
        
        Save $\langle s(t),a(t),r(t),p(t) \rangle$ in the replay buffer.
        
        Sample a mini-batch from the buffer.
        
        Update $\theta_1$  and $\theta_2$ every $O$ instants using~\eqref{DQN_update}.
        
        $t = t+1$.
    }
}

\caption{The proposed DRL algorithm with LSTM as the traffic predictor.}
\label{alg2}
\end{algorithm}

\begin{algorithm}[!t]
\SetAlgoLined
Define $\epsilon$, $\gamma$, $\alpha$ and $O$.

Define the reward function in \eqref{reward_function_2}

Initialize the replay buffer.

Define the number of episodes $E$.

\For{e = 1,...,$E$}{
    
    
    Choose a random value for $\zeta_1$ and $\zeta_2$ (action $a$) with probability $\epsilon$ or select the greedy action $a = \max_a Q(s(t),a)$ with probability $1-\epsilon$.
    
    Train the DQN in algorithm \ref{alg1} or \ref{alg2} to calculate the reward function.
    
    Save $\langle s(t),a(t),r(t),p(t) \rangle$ in the replay buffer.
    
    Sample a mini-batch from the buffer.
    
    Update $\theta_3$  and $\theta_4$ every $O$ instants using~\eqref{DQN_update}.
}

\caption{The reward function optimization.}
\label{alg3}
\end{algorithm}

\section{Key Performance Indicators}\label{EVAL_SEC}
In this section, we introduce the key performance indicators (KPIs). First, we present the KPIs related to the traffic estimation stage. Then, we discuss the KPIs of the DQNs in the UAV learning stage.

\subsection{Traffic Prediction KPIs}
\subsubsection{Mean square error}
The forward algorithm estimates recurrently the probability of a device to be active $\tilde{Y}_d$. One way to evaluate the estimation of the forward algorithm is to compare the resulting probability with the true activation probabilities $Y_d$. The mean square error (MSE) can be formulated as 
\begin{equation}
    MSE = \frac{1}{D} \sum_{d=1}^D(Y_d-\tilde{Y}_d)^2.
\end{equation}

\subsubsection{Training and validation losses of LSTM}
The loss quantifies the error in the prediction of machine learning models. A high loss indicates that the model generates an erroneous result, whereas a low loss indicates that the model is working well with few errors. The MSE loss function is the most well-known loss function in time-series regression-type problems~\cite{7860338}. The LSTM uses a sequence of data from the past (training data) to fit the weights of the gates. Herein, the training loss measures how well the model fits the training data. On the other hand, the estimated weights are used with new known data (validation data) to test how the optimized weights fit with new data in the future. Therefore, the validation loss measures how good the model is with future test data.

\subsubsection{LSTM classification metrics}
\begin{itemize}
    \item \textbf{The confusion matrix} presents the correct and wrong classification of each class in a matrix form.
    
    \item \textbf{The precision ($\mathcal{P}$)}is the ratio between the true predicted samples of a class and the total predicted samples of that class, whereas \textbf{the recall ($\mathcal{R}$)} is the ratio between the true predicted samples of a class and the total actual samples of that class. In addition, \textbf{the overall accuracy ($acc$)} is the ratio between the correct samples of both classes and the total samples.
    
    \item \textbf{The f1-score ($f1s$)} is calculated as follow
    \begin{equation}
    f1s = \frac{2 \: \mathcal{P}  \: \mathcal{R}}{\mathcal{P} + \mathcal{R}}.
\end{equation}
\end{itemize}

\subsection{UAV Learning KPIs}
\subsubsection{Immediate and accumulative reward}
In DRL models, an increasing immediate reward over the episodes is an important indication that the model learns. If the model has a decreasing immediate reward, this indicates that the learning scheme of the model is poor, whereas a fixed immediate reward over the episodes indicates the convergence of the DRL model and the possibility to terminate the training. The accumulative reward is an important evaluation metric to compare the DRL model with a baseline model such as the RW. In addition, it is an indicative KPI to compare multiple DRL with different hyperparameters, such as the learning rate, replay buffer size, and exploration rate, among others.

\subsubsection{Ergodic age}
The average age $\Bar{A}(t)$ at time instant $t$ is averaging the individual ages of each device. The ergodic age,
\begin{equation}
    A_e = \frac{1}{T} \sum_{t=1}^T\Bar{A}(t) = \frac{1}{T} \sum_{t=1}^T \frac{1}{D}\sum_{d=1}^D A_d(t),
\end{equation}
 is the mean of the average age over time. 


\subsubsection{Ergodic transmission power}
The average power $\Bar{P}(t)$ at time instant $t$ is the average of the individual powers of each device. The ergodic power is the time average of the accumulative power given as
\begin{equation}
    P_e = \frac{1}{T} \sum_{t=1}^T \Bar{P}(t) = \frac{1}{T} \sum_{t=1}^T\frac{1}{D}\sum_{d=1}^D P_d(t).
\end{equation}

\section{Simulation Results and Discussion}\label{RES_SEC}
In this section, we present the numerical results of the proposed DRL algorithm. First, we discuss the results of the traffic estimation via both the FA and the LSTM. Afterward, we exploit the proposed DQN to jointly optimize the AoI, regret, and transmission power. Finally, we present the results of optimizing the reward function for different network setups. We consider a grid world of  $11 \times 11$ cells, where each cell is a square with side length $100$ m. The simulation parameters are defined in Table~\ref{tab:uav}. We train the proposed algorithm using the Pytorch framework on a single NVIDIA Tesla V100 GPU and 20 GB of RAM. The RW scheme stands for random movement of the UAVs and random scheduling policy. The genie-aided scheme refers to the proposed DRL assuming perfect knowledge of the active and silent devices in the network. The term FA-DRL is used to describe the proposed DRL scheme with FA as the traffic predictor, whereas the term LSTM-DRL is used to describe the proposed DRL with LSTM as the traffic predictor.

\begin{table}[t!]
\centering
\caption{The UAV and DQN model parameters.}
\label{tab:uav}
\begin{tabular}{@{}cc@{}|cc@{}cc@{}}
\toprule
\textbf{Parameter}                                    & \textbf{Value} & \textbf{Parameter}                                    & \textbf{Value} \\ \midrule
$h_u$ & $100$ m & $h_{BS}$                          & $15$ m   \\
$A_{max}$  & 50  &  $L_g$ & 100 m \\

$E$                          & $10000$   & $e_{max}$                        & $200$        \\
                        
$B$                                        & $1 MHz$   & $M$                                       & $5$ Mb         \\

$\sigma^2$                                & $-100$ dBm   &  $v_u$                                       & $25$ m/s      \\

$v_{tip}$                                   & $120$ m/s   & $\rho$                                    & $1.225$ kg/m$^3$     \\

$P_0$                             & $99.66$ W &  $P_1$                          & $120.16$ W \\

 $d_0$                            & $0.48$    &  $\mu_0$                                  & $0.0001$    \\
 $Z$                                & $0.5$ s$^2$  &   $v_0$ & $0.002$ m/s    \\
 $g_0$   & $30$ dB    & $\alpha$   & $0.0004$  \\

$\epsilon$-decay & $0.995$  &  $\gamma$   & $0.99$ \\

DQN layers & $(64,128,128,64)$ & LSTM layers  & $(64,128,64)$ \\

Episodes  & $50000$ & Replay memory  & $100000$ \\

Activation fun. & ReLU & Optimizer & Adam \\

LSTM loss & Binary cross-entropy & DQN loss & MSE \\

\bottomrule
\end{tabular}
\end{table}

\begin{figure}[]
    \centering
    \subfloat[Error]{\includegraphics[width=0.51\textwidth,trim={0.1 0 0 0},clip]{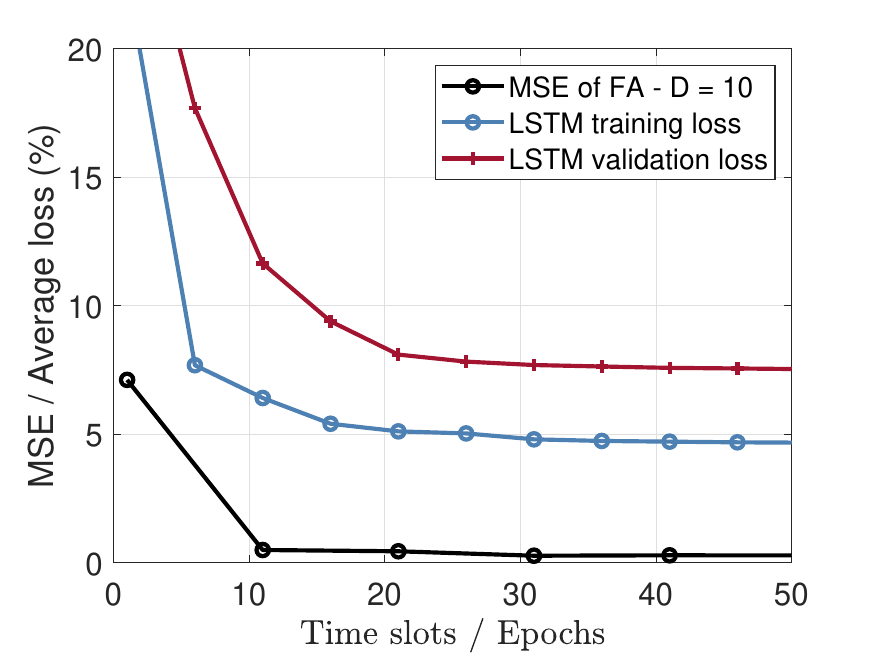}}
    \hskip -2.28ex
    \subfloat[Trajectory path]{\includegraphics[width=0.51\textwidth,trim={0.1 0 0 0},clip]{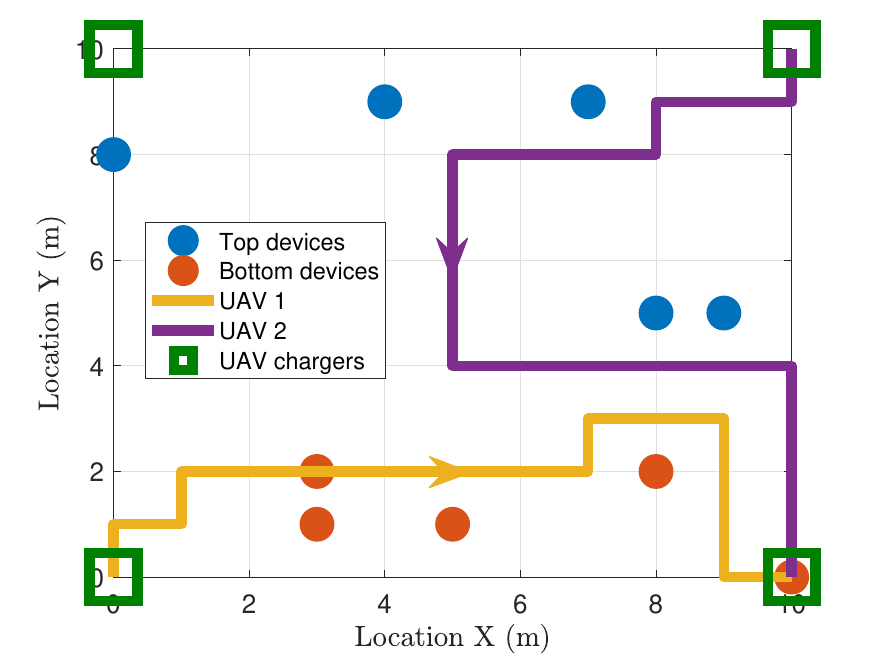}}
    \caption{(a) MSE of the FA activation probability prediction and training and validation losses of LSTM. (b) Trajectory path of 2-UAVs serving a network of $D = 10$ devices using the LSTM as the traffic predictor. The values for $\zeta_1$ and $\zeta_2$ are $25$ and $500$, respectively. The lower devices have a higher activation probability than the rest of the devices.} 
    \label{Fig4} 
\end{figure}


Fig.~\ref{Fig4}a depicts the MSE between the estimated activation probability using the FA and the true activation probability in two different network setups, namely, $D = 7$ and $D = 10$. We can notice that in the beginning, the error is high as the observations are not enough for the FA to estimate the hidden states and the future activation as discussed in~\eqref{activation_t_1}. Then, after 12-time slots, the MSE decreases to less than $0.5 \%$ and starts to converge. 
 Moreover, the average loss function is plotted versus epochs when using the LSTM to forecast the activation of the devices. In the beginning, the loss is higher as the weights of the LSTM architecture are randomly chosen. Afterward, the loss decreases as the weights are optimized using the accumulated device activation patterns. Convergence occurs after about 20 epochs where the training could be stopped.

Fig.~\ref{Fig4}b exploits the trajectory optimization result from the proposed DRL algorithm. Fig.~\ref{Fig4}b exploits the trajectory optimization result from the proposed DRL algorithm. For illustration, we present the trajectory optimization using LSTM as the traffic predictor. The devices at the bottom of the grid world are set to have higher activation probability by increasing the values $\epsilon^1$ and decreasing the values of $\epsilon^0$. This ensures a higher activation probability of the events that affect these devices. In addition, the values of $p_{dk}$ for the devices at the bottom of the network are higher than the other devices, forcing those devices to be active as long as possible. As shown in Fig.~\ref{Fig4}b, both UAVs tend to spend more time navigating near the bottom of the map. This indicates the effectiveness of the proposed learning scheme capturing that those devices are active most of the time. This trajectory is the optimized path that jointly minimizes age, regret, and transmission power.




\begin{figure*}[t]
    \centering
    \subfloat[Average age\label{Fig8a}]{\includegraphics[width=0.333\textwidth]{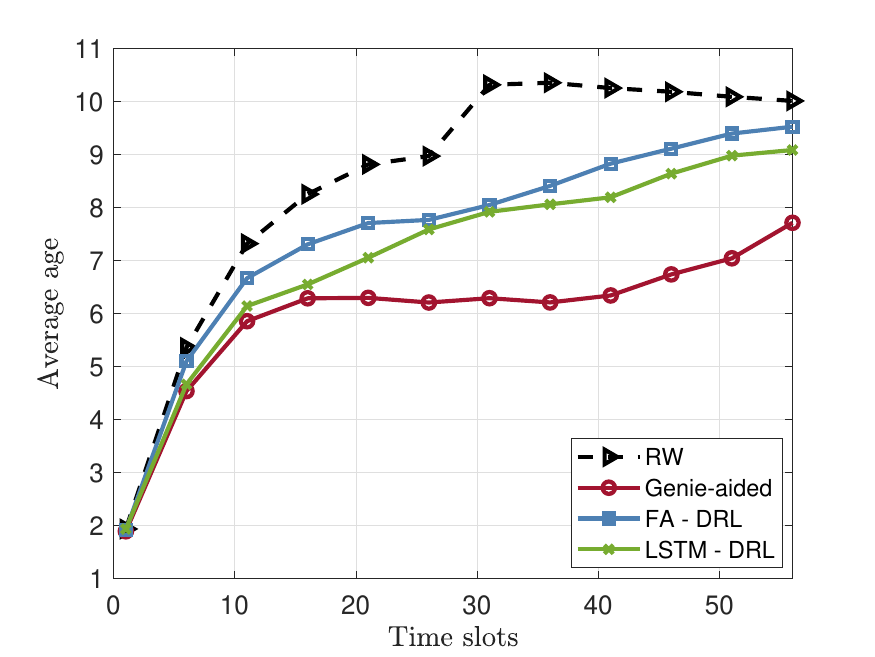}}
    \hskip -1.9ex
    \subfloat[Accumulative regret\label{Fig8b}]{\includegraphics[width=0.333\textwidth,trim={0 0 0 0},clip]{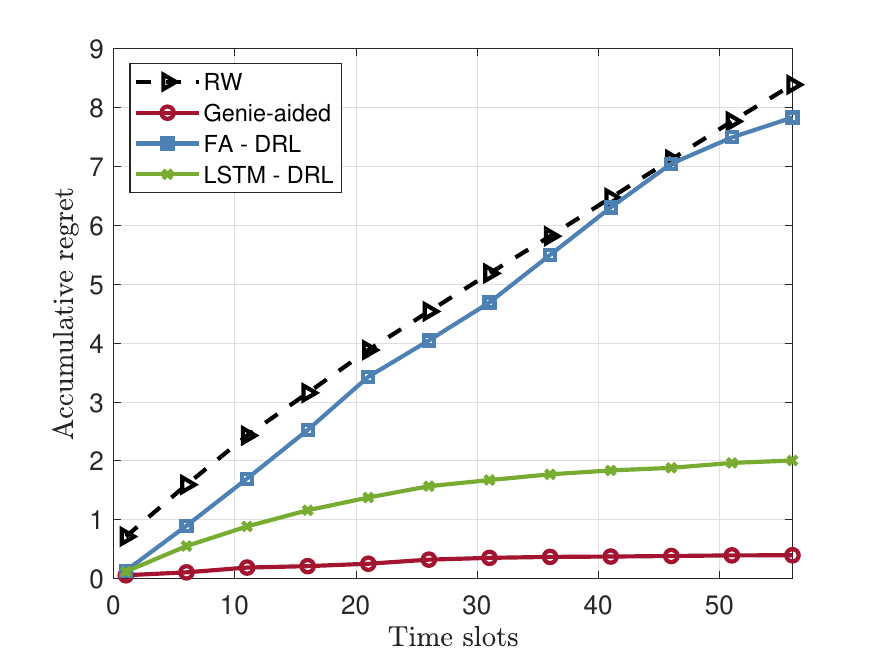}}
    \hskip -1.9ex
    \subfloat[Accumulative power\label{Fig8c}]{\includegraphics[width=0.333\textwidth,trim={0 0 0 0},clip]{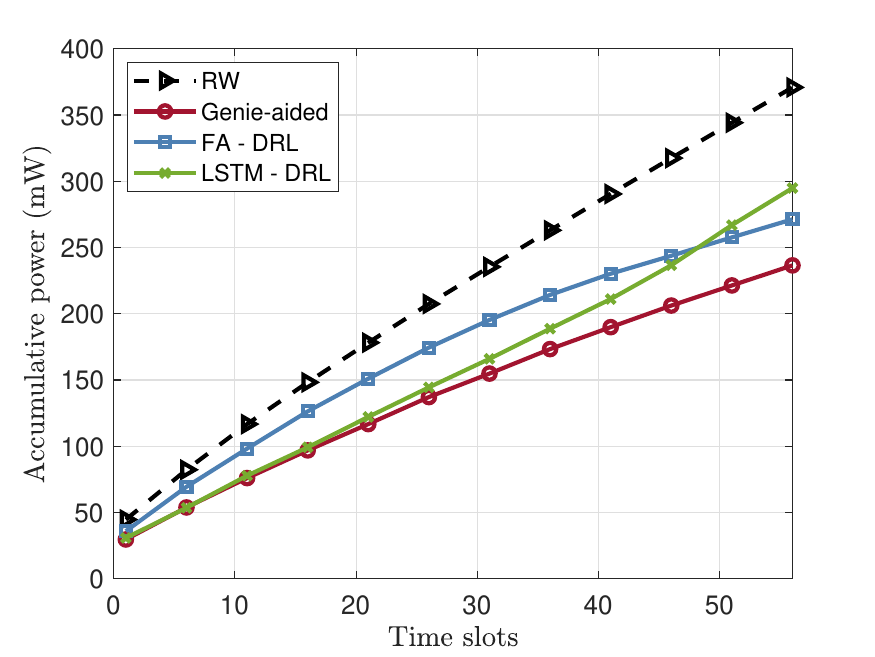}}
    \hskip -1.9ex
    \caption{The average age, accumulative regret, and accumulative power of a network of $D = 10$ devices served by two UAVs. The values for $\zeta_1$ and $\zeta_2$ are $25$ and $500$, respectively.}
    \label{Fig8} \vspace{-0mm}
\end{figure*}

\begin{figure}[t!]
    \centering  \vspace{-3mm}
    \subfloat[Ergodic age\label{Fig9a}]{\includegraphics[width=0.485
    \textwidth,trim={0.1 0 0 0},clip]{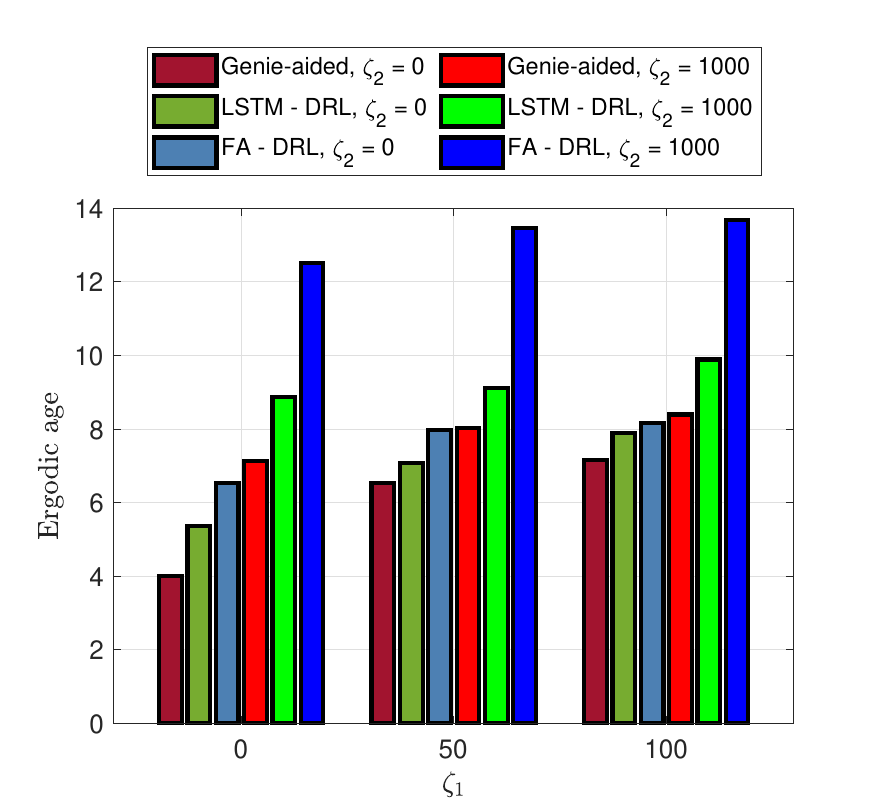}}
    \hskip -2.28ex
    \subfloat[Ergodic power\label{Fig9b}]{\includegraphics[width=0.485\textwidth,trim={0.1 0 0 0},clip]{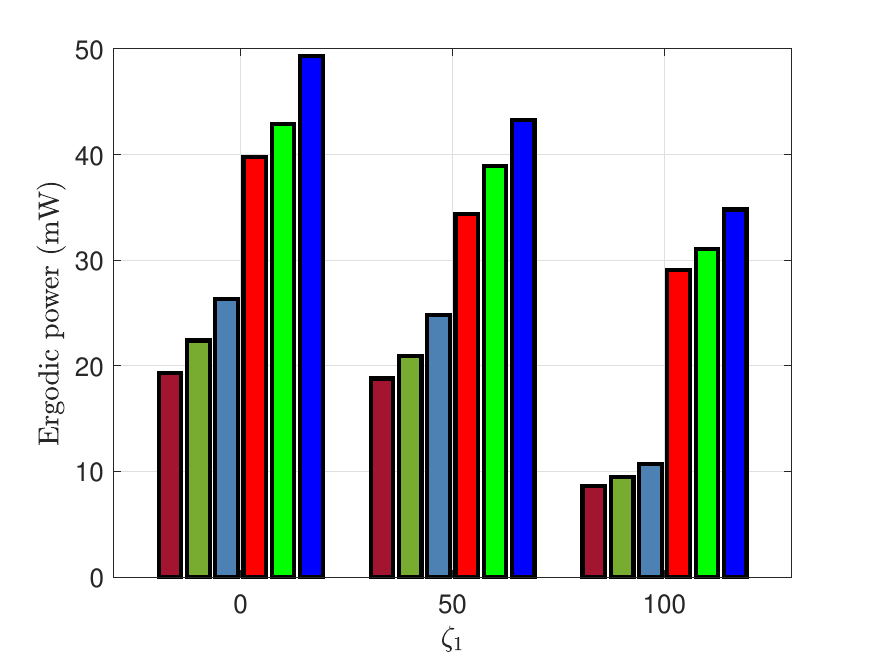}}
    \caption{The ergodic age and ergodic power of a network of $D = 10$ devices served by two UAVs while using different values of $\zeta_1$ and $\zeta_2$ using both the FA and LSTM as traffic predictors.}
    \label{Fig9} \vspace{-2mm}
\end{figure}

Table~\ref{tab:COMP} highlights comparing the FA and the LSTM as traffic predictors. The FA relies on the model parameters and prior observations to predict the probability of a device being active in future instants. On the other hand, the LSTM is model-free; it relies only on the previous observations to solve the activation of the devices. Therefore, both predictors require a long sequence of the previous activations to work efficiently. As the FA works recursively over all the previous time instants, including very long sequences to estimate the probability becomes cumbersome. This is not the case with the LSTM, which uses very long sequences efficiently to produce the actual activation pattern, thanks to the forget gate. We can notice in Table~\ref{tab:COMP} shows that the LSTM is more complex than the FA regarding time and memory consumption. 
However, if the BS uses the same long sequence for the FA as the LSTM, the FA becomes more complex than the LSTM regarding training time and memory consumption. We evaluate the FA performance using the MSE of activation probabilities. The FA has an average MSE of $0.0016$ for all devices. The LSTM returns the actual activation pattern (binary). Therefore, we evaluate the LSTM performance using the confusion matrix, where it has an average performance of correctly predicting $48$ active instants and $46$ silent instants from a total of $100$ time instants. This means that the LSTM predicts a device to be silent while it is active two times and it predicts a device to be active while it is silent four times. For the active instants, it has a precision of $ 92 \%$, recall of $ = 96 \%$, and $\mathbf{f1-score} = 94 \%$. Meanwhile, the silent instants have a precision of $ 96 \%$, recall of $ 92 \%$, and $\mathbf{f1-score} = 94 \%$. The LSTM has an overall accuracy of $94 \%$, which is quite high.



%

\begin{table}[t!]
\centering
\caption{A comparison of the performance of the FA and the LSTM as traffic predictors.}
\label{tab:COMP}
\begin{tabular}{@{}ccc@{}}
\toprule
\textbf{Parameter}                     & FA  & LSTM                                                                  \\ \midrule

Time complexity per device (training)  & $6 s$ & $150 s$  \\

DQN time complexity per episode (testing)  & $0.015 s$ & $0.2 s$  \\

Memory consumption per device & $190$ MB & $1200$ MB \\
Confusion matrix (100 instants) & $-$ & $\begin{pmatrix}
48 & 2 \\
4 & 46 \end{pmatrix}$ \\

Overall accuracy & $-$ & $94 \%$ \\

Mean square error & $0.16$ \% & $-$ \\

\bottomrule
\end{tabular}
\end{table}

Table~\ref{TABLE_III} demonstrates the immediate reward over episodes for the proposed DRL approach. It is noticeable that the reward enhances as more episodes are trained. This confirms that the DQN is learning over time. Fig.~\ref{Fig9} exploits the performance of the proposed algorithm using different reward functions, namely, different values for $\zeta_1$ and $\zeta_2$. It is noticeable that using $\zeta_1 = 0$ and $\zeta_2 = 0$ has the best performance concerning AoI. On the other hand, the accumulative regret and the accumulative power increase as they do not weigh the reward function. In addition, utilizing LSTM as the traffic predictor leads to lower AoI when compared to the FA. Using $\zeta_1 = 100$ and $\zeta_2 = 0$ increases the weight of the regret in the reward function, which results in the best accumulative regret using both proposed traffic predictors, whereas the average age and the accumulative power increase. Setting $\zeta_1 = 0$ and $\zeta_2 = 1000$ reduces the power consumption at the cost of worse AoI and regret, where both LSTM and the FA traffic predictors almost give the same accumulative power results.

\begin{table}[t!]
\centering
\caption{The reward function of training 2-UAVs serving a network of $D = 10$ devices using the LSTM as the traffic predictor. The values for $\zeta_1$ and $\zeta_2$ are $25$ and $500$, respectively.}
\label{TABLE_III}
\begin{tabular}{l|l|l|l|l|l|l}
\toprule
\textbf{Episode} & $1$ & $2000$ & $4000$ & $6000$ & $8000$ & $10000$ \\ \midrule
\textbf{Reward} & $-5700$ & $-4145$  &  $-1950$ & $-1660$ & $-1090$ & $-980$ \\
\bottomrule
\end{tabular}
\end{table}

In Fig.~\ref{Fig8}, the average AoI, the accumulative regret, and the accumulative transmission power are plotted over time for a network of $D=10$ devices served by two UAVs. The optimized values of $\zeta_1$ and $\zeta_2$ are $25$ and $500$, respectively. We can notice in Fig.~\ref{Fig8a} that the LSTM-DRL performs better concerning AoI than the FA-DRL, where both outperform the RW baseline scheme. In Fig.~\ref{Fig8b}, the performance of the FA-DRL is worse than the genie-aided and the LSTM-DRL due to the uncertainty of the traffic prediction that relies on the estimated probabilities. The accumulative power of the proposed FA-DRL and LSTM-DRL is lower than the RW scheme and almost catches the power of the genie-aided, as shown in Fig.~\ref{Fig8c}. Despite being more complex than the FA, the LSTM achieves better performance in terms of AoI, regret, and power consumption. Herein, we use the FA as a baseline model to be compared to the LSTM and show the trade-off between complexity and performance efficiency.

Fig.~\ref{Fig9} depicts the algorithm's ergodic age and ergodic power while sweeping the values of the power factor $\zeta_1$ on the x-axis. From Fig.~\ref{Fig9a}, we can notice that assigning $(0,0)$ to $(\zeta_1,\zeta_2)$  minimizes the ergodic AoI for all the DRL schemes. Assigning large values such as $(100,1000)$ to $(\zeta_1,\zeta_2)$ and  asymptotically up to $(\infty, \infty)$ render high AoI. On the other hand, Fig.~\ref{Fig9b} shows the effect of different values of $(\zeta_1,\zeta_2)$ on the ergodic power. We observe that high values of $\zeta_1$  minimize the transmission power and vice versa. Overall, The LSTM traffic predictor outperforms the FA traffic predictor regarding ergodic age and ergodic power consumption over time.

Note that, both Fig. \ref{Fig9a} and \ref{Fig9b} elucidate that the average AoI is directly proportional with the transmit power. This is because if the devices are able to transmit the signal with with hight power, this signal can travel to higher distances. Therefore, UAVs would receive the data without the need to move closer to the devices, which and updates arrive more frequently, which improves the AoI.

It is worth mentioning that, during training, the UAVs share their states with the central unit (BS), which sends back the actions and the rewards for each UAV. On the other hand, during testing, the UAVs save the look-up table of the converged state-action pairs by just exchanging the current states between the UAVs to avoid high signaling overhead.


\section{Conclusion}\label{CONC_SEC}
In this paper, we proposed a novel framework to jointly minimize the average AoI, regret, and transmission power of the IoT devices by optimizing the trajectory path of multiple UAVs and their scheduling policy. First, in the traffic estimation stage, the BS predicts the traffic of the IoT devices using a classical approach (FA) and a deep learning approach (LSTM). Then, we propose a DQN solution and select the optimum reward function in the UAV learning stage by optimizing the importance weights of the regret and the transmission power, \textit{i.e.}, $\zeta_1$ and $\zeta_2$, respectively. Finally, the optimal policy regarding the trajectory of the UAVs and their scheduling is optimized. The simulation results elucidate that the LSTM outperforms the FA in predicting the traffic of the devices to be used in the UAV learning stage. The LSTM has higher time and space complexity demands than the FA. Furthermore, the BS stage chooses the best reward function for the UAV learning stage. In the UAV learning stage, the proposed DQN approach shows better results regarding the AoI, regret, and transmission power than the baseline RW scheme.

Note that, in our endeavour, we considered only a static deployment of devices to illustrate the idea. In order to solve the same problem in a dynamic environment, meta-learning could be applied, where the experience of solving a specific setup could be invested in order to facilitate the learning process when devices move around. Finally, we would like to note that increasing the number of UAVs and devices are open research questions for future investigation using distributed learning approaches.



\bibliographystyle{IEEEtran}
\bibliography{references}
\end{document}